\documentclass[letterpaper,english,conference]{IEEEtran}
\usepackage[T1]{fontenc}
\usepackage[latin9]{inputenc}
\PassOptionsToPackage{algoruled, linesnumbered}{algorithm2e}
\usepackage{babel}
\usepackage{booktabs}
\usepackage{algorithm2e}
\usepackage{amsmath}
\usepackage{amssymb}
\usepackage{graphicx}
\usepackage[unicode=true]
 {hyperref}

\makeatletter

\pdfpageheight\paperheight
\pdfpagewidth\paperwidth

\providecommand{\tabularnewline}{\\}

\def\ps@IEEEtitlepagestyle{
  \def\@oddfoot{\mycopyrightnotice}
  \def\@evenfoot{}
}
\def\mycopyrightnotice{
  {\footnotesize $\copyright$ 2020 IEEE \hfill}
  \gdef\mycopyrightnotice{}
}

\makeatother

\begin{document}
\title{Learned Weight Sharing for Deep Multi-Task Learning by Natural Evolution
Strategy and Stochastic Gradient Descent}
\author{\IEEEauthorblockN{Jonas Prellberg}
\IEEEauthorblockA{\textit{Dept. of Computer Science} \\
\textit{University of Oldenburg}\\
Oldenburg, Germany \\
jonas.prellberg@uni-oldenburg.de}
\and
\IEEEauthorblockN{Oliver Kramer}
\IEEEauthorblockA{\textit{Dept. of Computer Science} \\
\textit{University of Oldenburg}\\
Oldenburg, Germany \\
oliver.kramer@uni-oldenburg.de}}
\maketitle
\begin{abstract}
In deep multi-task learning, weights of task-specific networks are
shared between tasks to improve performance on each single one. Since
the question, which weights to share between layers, is difficult
to answer, human-designed architectures often share everything but
a last task-specific layer. In many cases, this simplistic approach
severely limits performance. Instead, we propose an algorithm to \emph{learn}
the assignment between a shared set of weights and task-specific layers.
To optimize the non-differentiable assignment and at the same time
train the differentiable weights, learning takes place via a combination
of natural evolution strategy and stochastic gradient descent. The
end result are task-specific networks that share weights but allow
independent inference. They achieve lower test errors than baselines
and methods from literature on three multi-task learning datasets.
\end{abstract}

\section{Introduction}

Deep learning systems have achieved remarkable success in various
domains at the cost of massive amounts of labeled training data. This
poses a problem in cases where such data is difficult or costly to
acquire. In contrast, humans learn new tasks with minimal supervision
by building upon previously acquired knowledge and reusing it for
the new task. Transferring this ability to artificial learning is
a long-standing goal that is being tackled from different angles \cite{DBLP:conf/ijcai/LeeLKKKZ16,DBLP:journals/corr/FernandoBBZHRPW17,DBLP:conf/aaai/TesslerGZMM17}.
A step in this direction is multi-task learning (MTL), which refers
to learning multiple tasks at once with the intention to transfer
and reuse knowledge between tasks in order to better solve each single
task \cite{Caruana1997}.

MTL is a general concept that can be applied to learning with different
kinds of models. For the case of neural networks, MTL is implemented
by sharing some amount of weights between task-specific networks (hard
parameter sharing) or using additional loss functions or other constraints
to create dependencies between otherwise independent weights of task-specific
networks (soft parameter sharing). This way, overfitting is reduced
and better generalization may be achieved because the network is biased
to prefer solutions that apply to more than one task.

\begin{figure}
\begin{centering}
\includegraphics{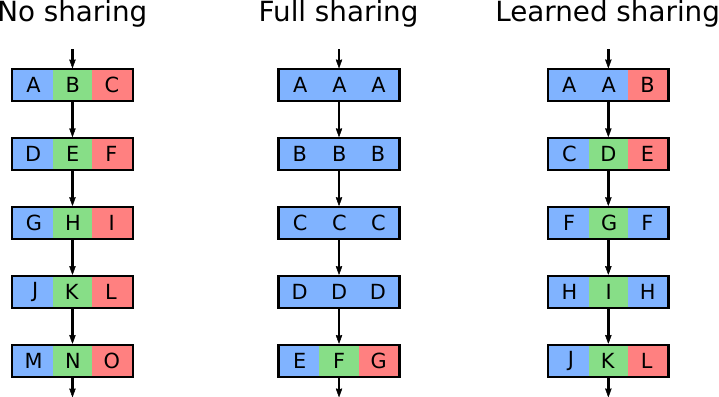}
\par\end{centering}
\caption{\label{fig:exposee}Different weight sharing schemes to solve a three-task
MTL problem. Each box depicts a network layer and the letters denote
different weights. Colored partitions in each box illustrate which
tasks share weights for that layer. Weights are never shared in the
last layer because it is the task-specific output layer. \emph{No
sharing:} When disregarding the possibility to perform MTL, an independent
set of weights is used in every task-specific network. \emph{Full
sharing:} Sometimes called shared back-bone in literature, this scheme
shares weights in all but the final layer. \emph{Learned sharing:}
A weight assignment learned by our method on the DKL-MNIST dataset.}
\end{figure}

Weight sharing can take many forms but for this paper we confine ourselves
to hard parameter sharing and the common case where weights can only
be shared between corresponding layers of each task-specific network.
Figure~\ref{fig:exposee} illustrates the resulting spectrum of possible
sharing configurations.

The difficulty then lies in choosing an appropriate weight sharing
configuration from this extremely large search space. We introduce
an automatic method that learns how to share layer weights between
task-specific networks using alternating optimization with a natural
evolution strategy (NES) and stochastic gradient descent (SGD). The
main problem is the non-differentiable assignment between weights
and layers that prevents learning both the assignment and weights
with SGD. Therefore, we exploit the black-box nature of NES to optimize
a probability distribution over the non-differentiable assignment.
It would also be possible to learn the weights themselves with NES
but this is rather inefficient compared to SGD. Since for every fixed
assignment the networks become differentiable wrt. their weights,
we exploit SGD to efficiently train them.

While alternating these two steps, the probability distribution's
entropy decreases and the layer weights are optimized to perform well
under the most likely assignments. In the end, this results in a single
most likely assignment and corresponding layer weights. Notably, this
is achieved without resorting to costly fitness evaluation steps that
have to train networks from scratch, or differentiable weight sharing
approaches \cite{DBLP:conf/cvpr/MisraSGH16,DBLP:journals/corr/RuderBAS17,DBLP:conf/iclr/MeyersonM18}
that result in computationally intensive forward passes during inference.

Using our learned weight sharing (LWS) method, we show accuracy improvements
compared to our own baselines and baselines from literature on three
datasets.

\section{Related Work}

The general approach that powers LWS is a hybrid optimization of differentiable
and non-differentiable parameters. Such a concept is used in \cite{DBLP:conf/icml/AkimotoSYUSN19}
to perform neural architecture search. The non-differentiable parameters
govern what kinds of layers are used in the architecture and are optimized
in an alternating fashion together with the layer weights. Similarly,
in \cite{DBLP:journals/corr/abs-1906-03139} the non-differentiable
parameters are sparsity masks inside a network and the differentiable
parameters are again layer weights. This allows to train sparse networks
directly instead of sparsifying them after first training a dense
model.

Next, we will present deep MTL literature to position our work against
existing methods and then NES literature to provide background for
understanding the method.

\subsection{Deep Multi-Task Learning}

The main difference between various deep multi-task learning approaches
is how weight sharing between tasks is implemented. This decision
is encoded in the architecture of a deep neural network, either by
the designer or an algorithm. Early works usually employ a shared
neural network that branches into small task-specific parts at its
end \cite{DBLP:conf/eccv/ZhangLLT14,DBLP:conf/naacl/LiuGHDDW15}.
This approach, referred to as full sharing in this paper, is restrictive
because all tasks have to work on exactly the same representation,
even if the tasks are very different. This motivated further work
to lift this restriction and make weight sharing data-dependent.

Approaches like cross-stitch networks \cite{DBLP:conf/cvpr/MisraSGH16},
Sluice networks \cite{DBLP:journals/corr/RuderBAS17} or soft layer
ordering \cite{DBLP:conf/iclr/MeyersonM18} introduce additional parameters
that control the weight sharing and are jointly optimized with the
networks weights by SGD. In these approaches, the task-specific networks
are connected by gates between every layer that perform weighted sums
between the individual layer's outputs. The coefficients of these
weighted sums are learned and can therefore control the influence
that different tasks have on each other. However, since all task-specific
networks are interconnected, this approach requires to evaluate all
of them even when performing inference on only a single task. In contrast,
every task-specific network in LWS can be used for inference independently.
Soft layer ordering further has the restriction that all shareable
layers at any position in the network must be compatible in input
and output shape. LWS on the other hand is unrestricted by the underlying
network architecture and can for example be applied to residual networks.

Another set of works explores non-differentiable ways to share weights.
Examples include fully-adaptive feature sharing, which iteratively
builds a branching architecture that groups similar tasks together
\cite{DBLP:conf/cvpr/LuKZCJF17} or routing networks, which use reinforcement
learning to choose a sequence of modules from a shared set of modules
in a task-specific way \cite{DBLP:conf/iclr/RosenbaumKR18}. Routing
networks are similar to our work in that they also avoid the interconnection
between task-specific networks as described before. However, their
approach fundamentally differs in that their routing network chooses
layers in a data-dependent way on a per-example basis, while our network
configuration is fixed after training and only differs between tasks
not examples.

\subsection{Natural Evolution Strategy}

Natural Evolution Strategy refers to a class of black-box optimization
algorithms that update a search distribution in the direction of higher
expected fitness using the natural gradient \cite{DBLP:journals/jmlr/WierstraSGSPS14}.
Given a parameterized search distribution with probability density
function $q\left(\left.x\right|\alpha\right)$, and a fitness function
$u\left(x\right)$, the expected fitness is
\begin{equation}
J\left(\alpha\right)=\mathbb{E}_{q_{\alpha}}\left[u\left(x\right)\right].
\end{equation}
The plain gradient in the direction of higher expected fitness can
be approximated from samples $x_{1},\ldots,x_{\lambda}$ distributed
according to $q\left(\left.x\right|\alpha\right)$ by a Monte-Carlo
estimate as
\begin{equation}
\nabla_{\alpha}J\left(\alpha\right)\approx\frac{1}{\lambda}\sum_{i=1}^{\lambda}u\left(x_{i}\right)\nabla_{\alpha}\log q\left(\left.x_{i}\right|\alpha\right)\label{eq:nes-gradient}
\end{equation}
with population size $\lambda$. Instead of following the plain gradient
directly, NES follows the natural gradient $\mathbf{F}^{-1}\nabla_{\alpha}J\left(\alpha\right)$.
Here, $\mathbf{F}^{-1}$ refers to the inverse of the search distribution's
Fisher information matrix. The natural gradient offers, among others,
increased convergence speed on flat parts of the fitness landscape.

The Fisher information matrix depends only on the probability distribution
itself and can often be analytically derived, e.g. for the common
case of multinormal search distributions \cite{DBLP:conf/icml/YiWSS09}.
For the case of search distributions from the exponential family under
expectation parameters, a direct derivation of the natural gradient
without first calculating $\mathbf{F}$ exists \cite[page 57]{DBLP:journals/jmlr/OllivierAAH17}.
The probability density function for members of the exponential family
has the form
\begin{equation}
q\left(\left.x\right|\alpha\right)=h\left(x\right)\exp\left\{ \alpha\cdot T\left(x\right)-A\left(\alpha\right)\right\} \label{eq:exp-family-density}
\end{equation}
with natural parameter vector $\alpha$, sufficient statistic vector
$T\left(x\right)$ and cumulant function $A\left(\alpha\right).$
We focus only on the case where $h\left(x\right)=1$ in our paper.
If we reparameterize the distribution with a parameter vector $\mu$
that satisfies
\begin{equation}
\mu=\mathbb{E}_{q_{\alpha}}\left[T\left(x\right)\right]=\nabla_{\alpha}A\left(\alpha\right),\label{eq:expectation-parameters}
\end{equation}
then we call $\mu$ the expectation parameters. With such a parameterization,
there is a nice result regarding the natural gradient: The natural
gradient wrt. the expectation parameters is given by the plain gradient
wrt. the natural parameters, i.e.
\begin{equation}
\tilde{\nabla}_{\mu}q\left(\left.x\right|\mu\right)=\nabla_{\alpha}q\left(\left.x\right|\alpha\right).
\end{equation}
It follows that the log-derivative, which is necessary for the search
gradient estimate in NES, can easily be derived as
\begin{align}
\tilde{\nabla}_{\mu}\log q\left(\left.x\right|\mu\right) & =\nabla_{\alpha}\log q\left(\left.x\right|\alpha\right)\\
 & =\nabla_{\alpha}\left(\alpha\cdot T\left(x\right)-A\left(\alpha\right)\right)\\
 & =T\left(x\right)-\mu\label{eq:natural-gradient}
\end{align}
because of the relationship in Equation~\ref{eq:expectation-parameters}
between gradient of the cumulant function and expectation parameters. 

In other words, if we choose a search distribution with expectation
parameters, the plain and natural gradient coincide. We will use this
fact later, to follow the natural gradient of a categorical distribution.

\section{Learned Weight Sharing}

\begin{figure*}[t]
\begin{centering}
\includegraphics{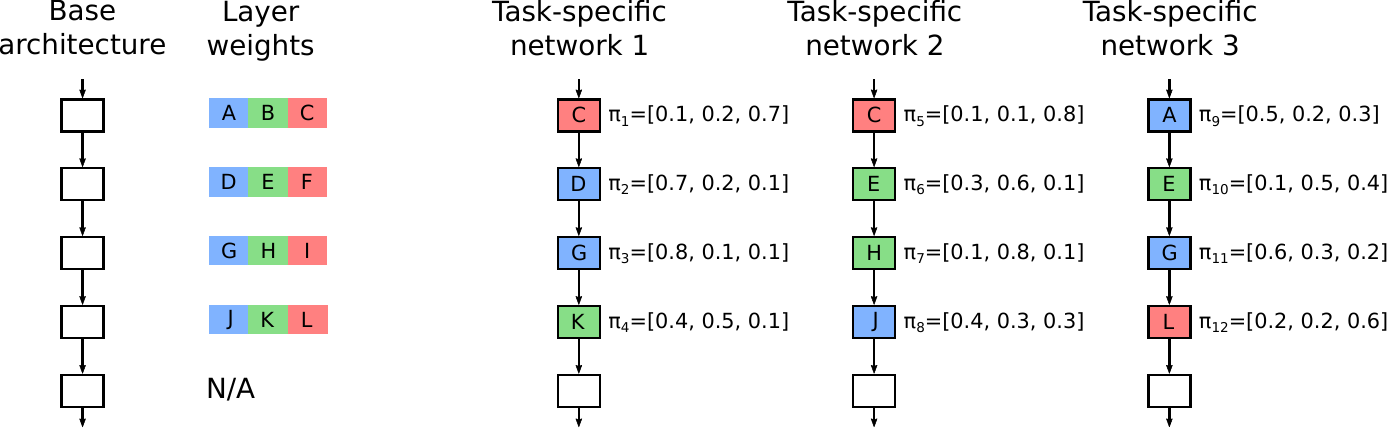}
\par\end{centering}
\caption{\label{fig:method}Setup to solve a three-task MTL problem with LWS.
A base architecture is duplicated for each task and weights are stochastically
assigned to each layer. There are a total of $N=12$ layers and $K=3$
weights per weight set. The depicted assignment is the most probable
one, which is used for inference.}
\end{figure*}

Consider the setup depicted in Figure~\ref{fig:method} to solve
an MTL problem using deep neural networks. Any neural network architecture
is chosen as the base architecture, e.g. a residual network. This
base architecture is duplicated once for each task to create task-specific
networks. Finally, the last layer of each task-specific network is
modified to have the appropriate number of outputs for the task.

In this setup, the weights of every layer except for the last one
are compatible between task-specific networks and can potentially
be shared. To this end, a set of $K$ weights is created for every
layer and all of the $N$ task-specific network layers are assigned
a weight from its corresponding set. By assigning the same weight
to multiple task-specific networks, weight sharing is achieved. The
number of weights per layer must not necessarily be the same, however
we restrict ourselves to equally sized sets of weights for simplicity.

The problem is now to find good assignments between the weights and
task-specific network layers and at at the same time train the weights
themselves. We achieve this by alternating between the optimization
of a search distribution over assignments with NES and the optimization
of layer weights with SGD. This approach is summarized in Algorithm~\ref{alg:lws}
and explained in more detail below.

\begin{algorithm}[t]
\DontPrintSemicolon
\SetProgSty{}
\newcommand{\myfuncsty}[1]{$\mathtt{#1}$}
\SetFuncSty{myfuncsty}
\SetProcArgSty{}
\SetKwProg{Def}{def}{ as}{end}
\SetKwFunction{FNES}{StepNES}
\SetKwFunction{FSGD}{StepSGD}
Let $p\left(\left.a\right|\pi\right)$ be the search distribution
over assignments\;
Let $f_{x,y}\left(\theta,a\right)$ be the loss for a batch of data
$x,y$ under weights $\theta$ and assignment $a$\;
\Def{\FNES$\left(\theta,\pi\right)$}{\label{lws:nes-start}
$x,y\leftarrow$ get random batch\;
\For{$i$ in $1 \ldots \lambda_\pi$}{
sample $a_{i}$ distributed according to $p\left(\left.a\right|\pi\right)$\;
calc. loss $l_{i}=f_{x,y}\left(\theta,a_{i}\right)$\;
calc. log-derivative $\nabla_{\pi}\log p\left(\left.a_{i}\right|\pi\right)$\label{lws:nes-log-derivative}\;
}
calc. utilities $u_{i}=2\cdot\frac{\textrm{rank}\left(l_{i}\right)-1}{\lambda_{\pi}-1}-1$\;
$\nabla_{\pi}J_{\pi}=\frac{1}{\lambda_{\pi}}\sum_{i=1}^{\lambda_{\pi}}u_{i}\nabla_{\pi}\log p\left(\left.a_{i}\right|\pi\right)$\;
\Return{$\pi+\eta_{\pi}\nabla_{\pi}J_{\pi}$}\;
}\label{lws:nes-end}
\Def{\FSGD$\left(\theta,\pi\right)$}{\label{lws:sgd-start}
$x,y\leftarrow$ get random batch\;
\For{$i$ in $1 \ldots \lambda_\theta$}{
sample $a_{i}$ distributed according to $p\left(\left.a\right|\pi\right)$\;
calc. weight gradient $\nabla_{\theta}f_{x,y}\left(\theta,a_{i}\right)$\;
}
$\nabla_{\theta}J_{\theta}=\frac{1}{\lambda_{\theta}}\sum_{i=1}^{\lambda_{\theta}}\nabla_{\theta}f_{x,y}\left(\theta,a_{i}\right)$\;
\Return{$\theta-\eta_{\theta}\nabla_{\theta}J_{\theta}$}\;
}\label{lws:sgd-end}
$\pi\leftarrow$ search distribution parameter vector with all $N\left(K-1\right)$
elements set to $\frac{1}{K}$\;
$\theta\leftarrow$ randomly initialized neural network weights\;
\While{not finished}{
$\pi\leftarrow$ \FNES$\left(\theta,\pi\right)$\;
$\theta\leftarrow$ \FSGD$\left(\theta,\pi\right)$\;
}
\caption{\label{alg:lws}LWS training procedure.}
\end{algorithm}

\subsection{Learning Objective}

The search for good assignments and layer weights is cast as an optimization
problem
\begin{equation}
\min_{\theta,a}f\left(\theta,a\right),
\end{equation}
where $f:\Theta\times\mathcal{A}\rightarrow\mathbb{R}$ is the average
loss over all tasks, $\theta\in\Theta$ is a vector of all layer weights,
and $a\in\mathcal{A}$ is an assignment of weights to task-specific
network layers. The loss function $f$ is differentiable wrt. $\theta$
but black-box wrt. $a$. We would like to exploit the fact that $\theta$
can be efficiently optimized by SGD but need a way to simultaneously
optimize $a$. Therefore, we create a stochastic version of the problem
\begin{equation}
\min_{\theta,\pi}J\left(\theta,\pi\right)=\mathbb{E}_{p_{\pi}}\left[f\left(\theta,a\right)\right]\label{eq:objective-J}
\end{equation}
by introducing a probability distribution defined on $\mathcal{A}$
with density function $p\left(\left.a\right|\pi\right)$. This stochastic
formulation makes the assignments amenable for optimization through
$\pi$ by the NES algorithm, but on the other hand requires to sample
assignments for the calculation of the gradient wrt. $\theta$.

\subsection{Assignment Optimization}

We use the NES algorithm to optimize $\pi$ for lower expected loss
$J\left(\theta,\pi\right)$ while keeping $\theta$ fixed (cf. Algorithm~\ref{alg:lws},
lines~\ref{lws:nes-start}~to~\ref{lws:nes-end}). The parameter
$\pi$ is initialized so that all assignments are equally probable
but with prior knowledge the initial parameter vector could also be
chosen so that it is biased towards certain preferred assignments.

Assignments $a_{1},\ldots,a_{\lambda_{\pi}}$ distributed according
to $p\left(\left.a\right|\pi\right)$ are sampled and their loss values
$l_{i}=f\left(\theta,a_{i}\right)$ are calculated on the same batch
of training data for all assignments. Following Equation~\ref{eq:nes-gradient},
the search gradient is approximated as
\begin{equation}
\nabla_{\pi}J\left(\theta,\pi\right)\approx\frac{1}{\lambda_{\pi}}\sum_{i=1}^{\lambda_{\pi}}u_{i}\nabla_{\pi}\log p\left(\left.a_{i}\right|\pi\right)\label{eq:gradient-pi}
\end{equation}
with utility values $u_{i}$ in place of fitness values (see below).
Finally, $\pi$ is updated by performing a step in the direction of
$\nabla_{\pi}J\left(\theta,\pi\right)$ scaled by a learning rate
parameter $\eta_{\pi}$. This is basic SGD but in principle more sophisticated
optimizers like SGD with momentum or Adam could be used for this update
step as well.

The utility values are created by fitness shaping to make the algorithm
invariant to the scale of the loss function. Loss values $l_{i}$
are transformed into utility values
\begin{equation}
u_{i}=2\cdot\frac{\textrm{rank}\left(l_{i}\right)-1}{\lambda_{\pi}-1}-1,
\end{equation}
where $\textrm{rank}\left(l_{i}\right)$ ranks the loss values from
1 to $\lambda_{\pi}$ in descending order, i.e. the smallest $l_{i}$
receives rank $\lambda_{\pi}$. This results in equally spaced utility
values in $\left[-1,1\right]$ with the lowest loss value receiving
a utility value of 1.

\subsection{Layer Weight Optimization}

While we can use backpropagation to efficiently determine the weight
gradient $\nabla_{\theta}f\left(\theta,a\right)$ with $a$ fixed,
determining $\nabla_{\theta}J\left(\theta,\pi\right)$ with $\pi$
fixed on the stochastic problem version is not possible directly.
Instead, we use a Monte-Carlo approximation to optimize $\theta$
for lower expected loss $J\left(\theta,\pi\right)$ while keeping
$\pi$ fixed (cf. Algorithm~\ref{alg:lws}, lines~\ref{lws:sgd-start}~to~\ref{lws:sgd-end}).

In the beginning, all layer weights $\theta$ are randomly initialized.
For the Monte-Carlo gradient estimation, assignments $a_{1},\ldots,a_{\lambda_{\theta}}$
distributed according to $p\left(\left.a\right|\pi\right)$ are sampled
and backpropagation is performed for each sample. The same batch of
training data is used for the backpropagation step throughout this
process for every assignment. The resulting gradients $\nabla_{\theta}f\left(\theta,a_{i}\right)$
are averaged over all assignments, so that the final gradient is given
by
\begin{equation}
\nabla_{\theta}J\left(\theta,\pi\right)\approx\frac{1}{\lambda_{\theta}}\sum_{i=1}^{\lambda_{\theta}}\nabla_{\theta}f\left(\theta,a_{i}\right).
\end{equation}
Using this gradient, $\theta$ is updated by SGD with learning rate
$\eta_{\theta}$ but, again, more sophisticated optimizers could be
employed instead.

\subsection{Natural Gradient}

The NES search gradient calculation in Equation~\ref{eq:gradient-pi}
actually follows the plain gradient instead of the natural gradient
unless we take care to use a specific parameterization for the search
distribution. As previously explained, the natural gradient and plain
gradient coincide when the distribution is a member of the exponential
family and has expectation parameters. In our problem setting, there
are a total of $N$ layers distributed over all task-specific networks
that need to be assigned a weight from $K$ possible choices from
the weight set corresponding to each layer. We can model this with
categorical distributions, which are part of the exponential family,
as follows.

First, consider a categorical distribution over $K$ categories with
samples $x\in\mathbb{N}_{K}$. It is well known \cite{Geyer2016},
that the categorical distribution can be written in exponential family
form (cf. Equation~\ref{eq:exp-family-density}) with natural parameters
$\alpha\in\mathbb{R}^{K-1}$ as
\begin{align}
p_{\textrm{nat}}\left(\left.x\right|\alpha\right) & =\exp\bigl\{\alpha\cdot T_{\textrm{nat}}\left(x\right)-A_{\textrm{nat}}\left(\alpha\right)\bigr\}\\
T_{\textrm{nat}}\left(x\right) & =\begin{pmatrix}\delta_{1,x} & \cdots & \delta_{K-1,x}\end{pmatrix}\\
A_{\textrm{nat}}\left(\alpha\right) & =\log\left(1+\sum_{i=1}^{K-1}e^{\alpha_{i}}\right),
\end{align}
where $\delta_{i,j}$ is the Kronecker delta function that is 1 if
$i=j$ and 0 otherwise. Our goal is to have this distribution in expectation
parameters so that we can use the results mentioned before for the
natural gradient calculation. We can reparameterize the distribution
as
\begin{align}
p_{\textrm{ex}}\left(\left.x\right|\mu\right) & =\exp\bigl\{ r_{\textrm{ex}}\left(\mu\right)\cdot T_{\textrm{ex}}\left(x\right)-A_{\textrm{ex}}\left(\mu\right)\bigr\}\\
r_{\textrm{ex}}\left(\mu\right) & =\begin{pmatrix}\log\frac{\mu_{1}}{\mu_{K}} & \cdots & \log\frac{\mu_{K-1}}{\mu_{K}}\end{pmatrix}\\
T_{\textrm{ex}}\left(x\right) & =\begin{pmatrix}\delta_{1,x} & \cdots & \delta_{K-1,x}\end{pmatrix}\\
A_{\textrm{ex}}\left(\mu\right) & =-\log\mu_{K},
\end{align}
which gives us a parameter vector $\mu\in\left[0,1\right]^{K-1}$
with entries corresponding to the probabilities of all but the last
category. For notational convenience, we use $\mu_{K}=1-\sum_{i=1}^{K-1}\mu_{i}$
even though it is not technically part of the parameter vector.

To see that $\mu$ are expectation parameters, we compare it to the
derivative of the cumulant function in natural parameters (cf. Equation~\ref{eq:expectation-parameters}).
By using the relationship $\alpha_{i}=\log\frac{\mu_{i}}{\mu_{K}}$
it is easy to show that
\begin{align}
\frac{\partial A_{\textrm{nat}}\left(\alpha\right)}{\partial\alpha_{i}} & =\frac{e^{\alpha_{i}}}{1+\sum_{j=1}^{K-1}e^{\alpha_{j}}}=\mu_{i}
\end{align}
holds for all $i\in\mathbb{N}_{K-1}$.

Now, consider a joint of $N$ independent but \emph{not} identically
distributed categorical distributions with samples $a$ and parameters
$\pi$ so that
\begin{align}
a & =\begin{pmatrix}a_{1} & \cdots & a_{N}\end{pmatrix}\in\mathbb{N}_{K}^{N}\\
\pi & =\begin{pmatrix}\pi_{1} & \cdots & \pi_{N}\end{pmatrix}\in\left[0,1\right]^{N\left(K-1\right)}
\end{align}
are the concatenations of the samples and expectation parameters of
all $N$ categorical distributions, i.e. $\pi$ is the concatenation
of $N$ expectation parameter vectors $\pi_{i}\in\left[0,1\right]^{K-1}$.

Due to the independence of the $N$ categorical distributions, the
density function for the joint distribution becomes the product of
their individual densities. Again, this is a member of the exponential
family with expectation parameters:
\begin{align}
p\left(\left.a\right|\pi\right) & ={\textstyle \prod\nolimits}_{i=1}^{N}p_{\textrm{ex}}\left(\left.a_{i}\right|\pi_{i}\right)\label{eq:joint-search-distribution}\\
 & =\exp\bigl\{ r\left(\pi\right)\cdot T\left(a\right)-A\left(\pi\right)\bigr\}\\
r\left(\pi\right) & =\begin{pmatrix}r_{\textrm{ex}}\left(\pi_{1}\right) & \cdots & r_{\textrm{ex}}\left(\pi_{N}\right)\end{pmatrix}\\
T\left(a\right) & =\begin{pmatrix}T_{\textrm{ex}}\left(a_{1}\right) & \cdots & T_{\textrm{ex}}\left(a_{N}\right)\end{pmatrix}\\
A\left(\pi\right) & ={\textstyle \sum\nolimits}_{i=1}^{N}A_{\textrm{ex}}\left(\pi_{i}\right).
\end{align}

In summary, LWS uses $p\left(\left.a\right|\pi\right)$ from Equation~\ref{eq:joint-search-distribution}
as the density for its search distribution. The parameters $\pi$
are the concatenation of all but the last probabilities for each categorical
distribution. Since $\pi$ are expectation parameters, we can use
Equation~\ref{eq:natural-gradient} to calculate the natural gradient
as
\begin{equation}
\nabla_{\pi}\log p\left(\left.a\right|\pi\right)=T\left(a\right)-\pi
\end{equation}
and plug it into Algorithm~\ref{alg:lws} at line~\ref{lws:nes-log-derivative}.

\subsection{Inference}

After training has finished, the most likely weight assignment $\arg\max_{a}p\left(\left.a\right|\pi\right)$
is used for inference.

\section{Experiments}

\begin{table}
\caption{\label{tab:own-vs-baselines}Test error of learned weight sharing
compared to full sharing and no sharing baselines.}

\centering{}%
\begin{tabular}{lrrr}
\toprule 
 & DLK-MNIST & CIFAR-100 & Omniglot\tabularnewline
Method & ConvNet & ResNet18 & ResNet18\tabularnewline
\midrule
\midrule 
Full sharing & 14.16 $\pm$ 0.37 & 31.80 $\pm$ 0.44 & 10.97 $\pm$ 0.60\tabularnewline
No sharing & 12.80 $\pm$ 0.16 & 32.53 $\pm$ 0.32 & 15.82 $\pm$ 1.02\tabularnewline
Learned sharing & \textbf{11.83 $\pm$ 0.51} & \textbf{30.84 $\pm$ 0.49} & \textbf{10.70 $\pm$ 0.62}\tabularnewline
\bottomrule
\end{tabular}
\end{table}

We demonstrate the performance of LWS on three different multi-task
datasets using convolutional network architectures taken from other
MTL publications to compare our results to theirs. We also perform
experiments using a residual network \cite{DBLP:conf/cvpr/HeZRS16}
architecture to show applicability of LWS to modern architectures.
Furthermore, we provide two baseline results for all experiments which
are full sharing, i.e. every task shares weights with every other
task at each layer except for the last one, and no sharing, i.e. all
task-specific networks are completely independent. Note that a completely
independent network for each task means that its whole capacity is
available to learn a single task, whereas the full sharing network
has to learn all tasks using the same capacity. Depending on network
capacity, task difficulty and task compatibility we will see no sharing
outperform full sharing and also the other way around.

All experiments are repeated 10 times and reported with mean and standard
deviation. For statistical significance tests, we perform a one-sided
Mann-Whitney U test. The search distribution parameters $\pi$ are
initialized to $\frac{1}{K}$ so that layers are chosen uniformly
at random in the beginning. Furthermore, to prevent that a layer will
never be chosen again once its probability reaches zero, every entry
in $\pi$ is clamped above 0.1\,\% after the update step and then
$\pi$ is renormalized to sum to one. The layer weights $\theta$
are initialized with uniform He initialization \cite{DBLP:conf/iccv/HeZRS15}
and update steps on $\theta$ are performed with the Adam \cite{DBLP:journals/corr/KingmaB14}
optimizer. All images used in the experiments are normalized to $\left[0,1\right]$
and batches are created by sampling 16 training examples from each
different task and concatenating them. Full sharing, no sharing, and
LWS all use the same equal loss weighting between different tasks.
MTL is usually sensitive to this weighting and further improvements
might be achieved but its optimization is left for future work. All
source code is publicly available online\footnote{\href{https://github.com/jprellberg/learned-weight-sharing}{https://github.com/jprellberg/learned-weight-sharing}}.

\subsection{DKL-MNIST}

DKL-MNIST is a custom MTL dataset created from the Extended-MNIST
\cite{DBLP:conf/ijcnn/CohenATS17} and Kuzushiji-MNIST \cite{DBLP:journals/corr/abs-1812-01718}
image classification datasets. We select 500 training examples of
digits, letters and kuzushiji each for a total of 1,500 training examples
and keep the complete test sets for a total of 70,800 test examples.
Using only a few training examples per task creates a situation where
sharing features between tasks should improve performance. Since all
three underlying datasets are MNIST variants, the training examples
are $28\times28$ grayscale images but there are 10 digit classes,
26 letter classes and 10 kuzushiji classes in each task respectively.

For this small dataset, we use a custom convolutional network architecture
that consists of three convolutional layers and two dense layers.
The convolutions all have 32 filters, kernel size $3\times3$ and
are followed by batch normalization, ReLU activation and $2\times2$
max-pooling. The first dense layer has 128 units and is followed by
a ReLU activation, while the second dense layer has as many units
as there are classes for the task. LWS is applied to the three convolutional
layers and the first dense layer, i.e. the whole network except for
the task-specific last layer.

We train LWS and the two baselines for 5,000 iterations on DKL-MNIST
using a SGD learning rate of $\eta_{\theta}=10^{-3}$. Furthermore,
LWS uses $\lambda_{\theta}=\lambda_{\pi}=8$ samples for both SGD
and NES, and a NES learning rate of $\eta_{\pi}=10^{-2}$ to learn
to share sets of $K=3$ weights for each layer. We see in Table~\ref{tab:own-vs-baselines}
that full sharing performs worse than no sharing, i.e. there is negative
transfer when using the simple approach of sharing all but the last
layer. However, using LWS we find an assignment that is significantly
($p<0.01$) better than the no sharing baseline for a total error
of 11.83\,\%.

\subsection{CIFAR-100}

\begin{table}
\caption{\label{tab:own-vs-routing}Comparison against results from \cite{DBLP:conf/iclr/RosenbaumKR18}
using their network architecture.}

\centering{}%
\begin{tabular}{lr}
\toprule 
Method & CIFAR-100 Test Error {[}\%{]}\tabularnewline
\midrule
\midrule 
Cross-stitch networks \cite{DBLP:conf/iclr/RosenbaumKR18} & 47\tabularnewline
Routing networks \cite{DBLP:conf/iclr/RosenbaumKR18} & 40\tabularnewline
\midrule 
Full sharing & 39.08 $\pm$ 0.36\tabularnewline
No sharing & \textbf{36.50 $\pm$ 0.43}\tabularnewline
Learned sharing & 37.43 $\pm$ 0.53\tabularnewline
\bottomrule
\end{tabular}
\end{table}

The CIFAR-100 image classification dataset is cast as an MTL problem
by grouping the different classes into tasks by the 20 coarse labels
that CIFAR-100 provides. Each task then contains 5 classes and 2,500
training examples (500 per class) for a total of 50,000 training examples
and 10,000 test examples, all of which are $32\times32$ pixel RGB
images.

We employ the neural network architecture given by \cite{DBLP:conf/iclr/RosenbaumKR18}
to allow for a comparison against their results. It consists of four
convolutional layers and four dense layers. The convolutions all have
32 filters and kernel size $3\times3$, and are followed by batch
normalization, ReLU activation and $2\times2$ max-pooling. The first
three dense layers all have 128 units and are followed by a ReLU activation,
while the last dense layer has as many units as there are classes
for the task, i.e. 5 for all tasks on this dataset. In \cite{DBLP:conf/iclr/RosenbaumKR18}
they only apply their MTL method to the three dense layers with 128
units so we do the same for a fair comparison. This means the convolutional
layers always share their weights between all tasks.

We train LWS and the two baselines for 4,000 iterations on CIFAR-100
using a SGD learning rate of $\eta_{\theta}=10^{-3}$. Furthermore,
LWS uses $\lambda_{\theta}=\lambda_{\pi}=8$ samples for SGD and NES,
and a NES learning rate of $\eta_{\pi}=10^{-1}$ to learn to share
sets of $K=20$ weights for each layer. Table~\ref{tab:own-vs-routing}
shows that LWS, with a test error of 37.43\,\%, outperforms both
cross-stitch networks at 47\,\% test error and routing networks at
40\,\% test error. However, no sharing achieves even better results.
This can be attributed to the network capacity being small in relation
to the dataset difficulty. In this case, having 20 times more weights
is more important than sharing data between tasks.

\begin{figure}
\begin{centering}
\includegraphics[width=0.4\textwidth]{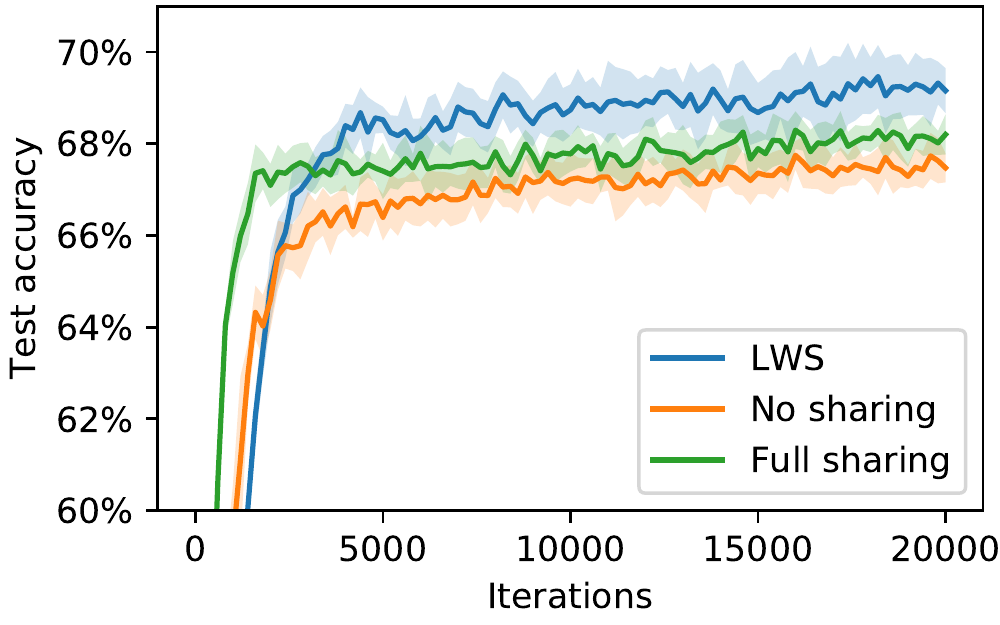}
\par\end{centering}
\caption{\label{fig:testacc-cifar100}Test accuracy during training for LWS
and its baselines using a ResNet18 on CIFAR-100.}
\end{figure}

Therefore, we repeat the experiment with a ResNet18 architecture that
has much higher capacity than the custom convolutional network from
\cite{DBLP:conf/iclr/RosenbaumKR18}. The channel configuration in
our ResNet18 is the same as in the original publication \cite{DBLP:conf/cvpr/HeZRS16}.
However, due to the much smaller image size of CIFAR-100, we remove
the $3\times3$ max-pooling layer and set the convolutional stride
parameters so that downsampling is only performed in the last three
stages. We apply LWS to share weights between each residual block.
They are treated as a single unit that consists of two convolutional
layers and, in the case of a downsampling block, a third convolutional
layer in the shortcut connection. All hyperparameters stay the same
except for the amount of iterations, which is increased to 20,000.
Test curves are shown in Figure~\ref{fig:testacc-cifar100} and final
test results are listed in Table~\ref{tab:own-vs-baselines}. We
notice that no sharing at 32.53\,\% test error now performs worse
than full sharing at 31.80\,\% test error. We believe the reason
to be the increased network capacity that is now high enough to benefit
from data sharing between tasks. LWS further improves on this and
achieves the lowest test error at 30.84\,\%, which is significantly
($p<0.01$) better than full sharing.

Depending on the sharing configuration, the total number of weights
that are present in the system comprised of all task-specific networks
differs. Naturally, the no sharing configuration has the highest possible
amount of weights at 223M, while full sharing has the lowest possible
amount at 11M. They differ exactly by a factor of 20, which is the
number of tasks in this setting. LWS finds a configuration that uses
136M weights while still achieving higher accuracy than both baselines.

\subsection{Omniglot}

\begin{table}
\caption{\label{tab:own-vs-slo}Comparison against results from \cite{DBLP:conf/iclr/MeyersonM18}
using their network architecture.}

\centering{}%
\begin{tabular}{lr}
\toprule 
Method & Omniglot Test Error {[}\%{]}\tabularnewline
\midrule
\midrule 
Soft layer ordering \cite{DBLP:conf/iclr/MeyersonM18} & 24.1\tabularnewline
\midrule 
Full sharing & 20.85 $\pm$ 1.07\tabularnewline
No sharing & 23.52 $\pm$ 1.25\tabularnewline
Learned sharing & \textbf{19.31 $\pm$ 2.54}\tabularnewline
\bottomrule
\end{tabular}
\end{table}

The Omniglot dataset \cite{Lake2015} is a standard MTL dataset that
consists of handwritten characters from 50 different alphabets, each
of which poses a character classification task. The alphabets contain
varying numbers of characters, i.e. classes, with 20 grayscale example
images of $105\times105$ pixels each. Since Omniglot contains no
predefined train-test-split, we randomly split off 20\,\% as test
examples from each alphabet.

We employ the neural network architecture given by \cite{DBLP:conf/iclr/MeyersonM18}
to allow for a comparison against their results. It consists of four
convolutional layers and a single dense layer. The convolutional layers
all have 53 filters and kernel size $3\times3$ and are followed by
batch normalization, ReLU activation and $2\times2$ max-pooling.
The final dense layer has as many units as there are classes for the
task. As in \cite{DBLP:conf/iclr/MeyersonM18}, we apply LWS to the
four convolutional layers.

We train LWS and the two baselines for 20,000 iterations on Omniglot
using a SGD learning rate of $\eta_{\theta}=10^{-3}$. Furthermore,
LWS uses $\lambda_{\theta}=\lambda_{\pi}=8$ samples for SGD and NES,
and a NES learning rate of $\eta_{\pi}=10^{-2}$ to learn to share
sets of $K=20$ weights for each layer. Table~\ref{tab:own-vs-slo}
shows how LWS outperforms SLO and both baselines. We repeat the experiment
with a ResNet18 architecture with the $3\times3$ max-pooling removed
and present results in Table~\ref{tab:own-vs-baselines}. LWS still
performs significantly ($p<0.01$) better than no sharing and is on
par with full sharing. Neither full sharing nor LWS is significantly
($p<0.01$) better than the other.

\subsection{Qualitative Results}

\begin{figure}
\begin{centering}
\includegraphics[width=0.33\textwidth]{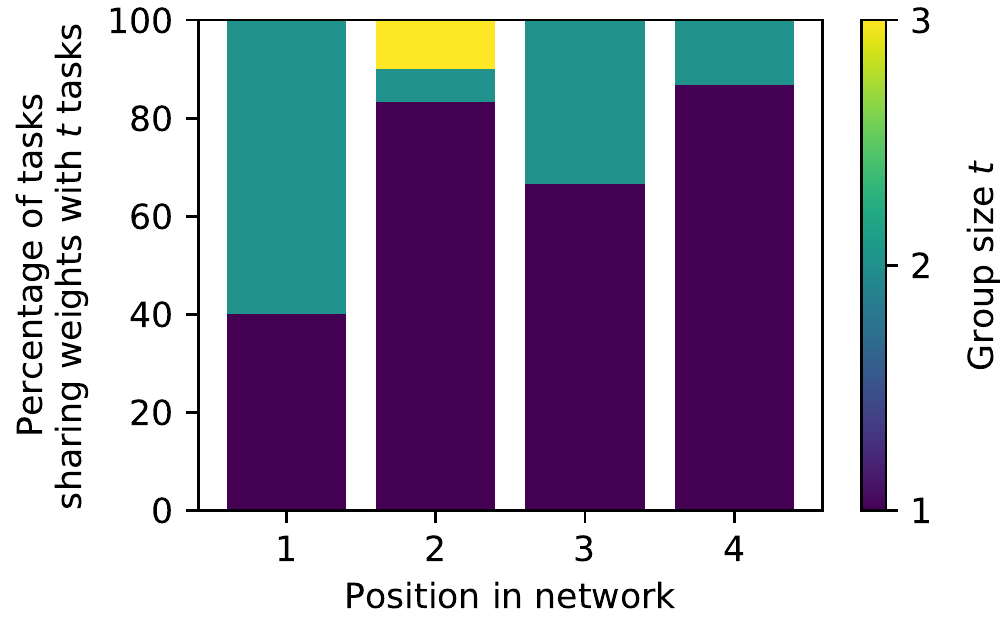}
\par\end{centering}
\caption{\label{fig:avg-sharing-mnist-convnet}Percentage of tasks that share
weights between exactly $t$ tasks when learned with LWS on DKL-MNIST.}
\end{figure}

Figure~\ref{fig:avg-sharing-mnist-convnet} sheds light on what kind
of assignments are learned on the DKL-MNIST dataset. The three convolutional
layers and the one dense layers that are shareable are denoted on
the horizontal axis in the same order as in the network itself. For
each layer, a stacked bar represents the percentage of tasks over
all repetitions that shared the layer weight within a group of $t$
tasks. Since there are three tasks and three weights per shared set
the only possible assignments are (1) all three tasks have independent
weights, (2) two tasks share the same weight, while the last task
has an independent weight, and (3) all three tasks share the same
weight. In Figure~\ref{fig:avg-sharing-mnist-convnet} the group
sizes correspond to these three assignments, e.g. in 40\,\% of the
experiments the first layer had three tasks with independent weights.
An exemplary assignment that was found in one of the DKL-MNIST experiments
can be seen in Figure~\ref{fig:exposee}.

\begin{figure}
\begin{centering}
\includegraphics[width=0.33\textwidth]{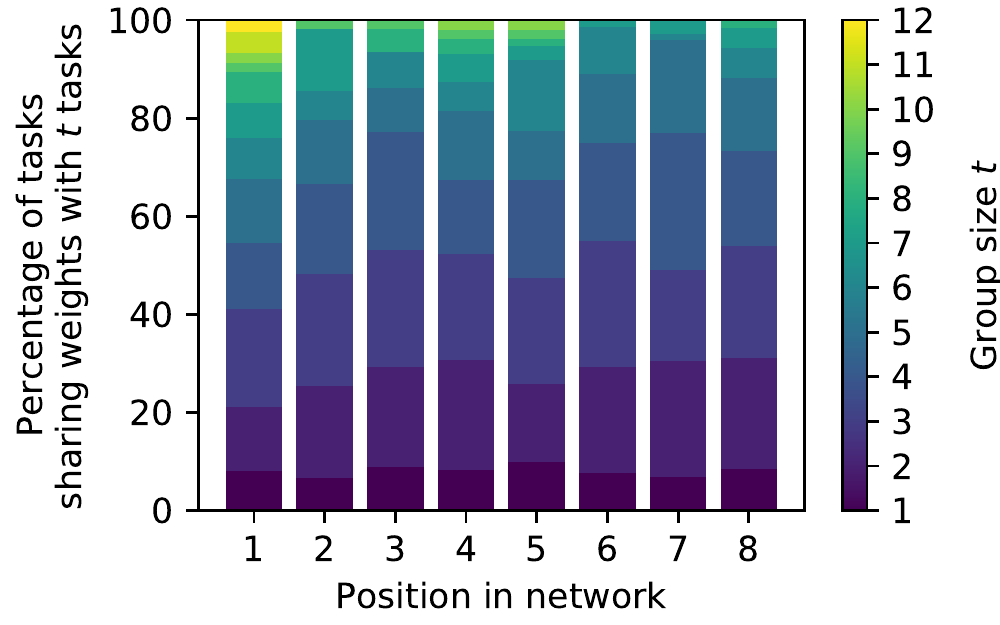}
\par\end{centering}
\caption{\label{fig:avg-sharing-omniglot-resnet}Percentage of tasks that share
weights between exactly $t$ tasks when learned with LWS on Omniglot.}
\end{figure}

Figure~\ref{fig:avg-sharing-omniglot-resnet} shows the same kind
of visualization on Omniglot for a ResNet18. Due to the vastly increased
number of possible assignments, the interpretation is not as straightforward
as in the DKL-MNIST case. However, we can clearly see how weights
are shared between a larger number of tasks in the early layers. This
corresponds well to results from transfer learning literature \cite{DBLP:conf/nips/YosinskiCBL14},
where early convolutional layers have been found to learn very general
filters.

\section{Conclusion}

LWS solves MTL problems by learning how to share weights between task-specific
networks. We show how combining NES and SGD creates a learning algorithm
that deals with the problem's non-differentiable structure by NES,
while still exploiting the parts that \emph{are} differentiable with
SGD. This approach beats the MTL approaches cross-stitch networks,
routing networks and soft layer ordering on their respective problems
and we show good performance on three datasets using a large-scale
residual network.

\bibliographystyle{IEEEtran}
\bibliography{lws}

\end{document}